\RequirePackage{fix-cm}
\documentclass[twocolumn]{svjour3}          
\smartqed  
\usepackage{graphicx}
\usepackage{url}
\usepackage[dvipdfmx]{hyperref}
\begin{document}

\title{cvpaper.challenge in 2015}
\subtitle{A review of CVPR2015 and DeepSurvey}


\author{Hirokatsu Kataoka \and
        Yudai Miyashita \and
		Tomoaki Yamabe \and
		Soma Shirakabe \and
		Shin'ichi Sato \and
		Hironori Hoshino \and
		Ryo Kato \and
		Kaori Abe \and
		Takaaki Imanari \and
		Naomichi Kobayashi \and
		Shinichiro Morita \and
		Akio Nakamura
}

\institute{Hirokatsu Kataoka \at
              Tsukuba, Ibaraki, Japan \\
              Tel.: +81-29-861-2267\\
              \email{hirokatsu.kataoka@aist.go.jp}
}

\date{Received: date / Accepted: date}

\maketitle

\begin{abstract}
The ``cvpaper.challenge" is a group composed of members from AIST, Tokyo Denki Univ. (TDU), and Univ. of Tsukuba that aims to systematically summarize papers on computer vision, pattern recognition, and related fields. For this particular review, we focused on reading the ALL 602 conference papers presented at the CVPR2015, the premier annual computer vision event held in June 2015, in order to grasp the trends in the field. Further, we are proposing ``DeepSurvey" as a mechanism embodying the entire process from the reading through all the papers, the generation of ideas, and to the writing of paper.
\end{abstract}

\section{Introduction}
\label{intro}
\textit{cvpaper.challenge} is a joint project aimed at reading papers mainly in the field of computer vision and pattern recognition~\footnote{Further reading: \href{https://twitter.com/cvpaperchalleng}{Twitter @CVPaperChalleng (https://twitter.com/cvpaperchalleng)}, \href{http://www.slideshare.net/cvpaperchallenge}{SlideShare @cvpaper.challenge (http://www.slideshare.net/cvpaperchallenge)}}. Currently the project is run by around ten members representing different organizations; namely, AIST, TDU, and University of Tsukuba~\footnote{In 2016, we are now around 30 members including the University of Tokyo and Keio University.}. Reading international conference papers clearly provides various advantages other than gaining an understanding of the current standing of your own research, such as acquiring ideas and methods used by researchers around the world. In reality, however, although this input of knowledge is important, researchers and engineers are too busy to have time to do it, and the process takes a great amount of time and effort for undergraduate and graduate students (particularly master’s course students) who lack research experience and entails sacrificing their time for classes and research. Assigning this work, however, to non-experts who are not familiar with the field of computer vision, results in a great amount of time needed for interpreting the papers. As a way to address this problem, we believe that we can make it relatively easier to grasp advanced technologies if we share and systematize knowledge using the Japanese language. We therefore undertook to extensively read papers, summarize them, and share them with others working in the same field. The IEEE-sponsored Conference on Computer Vision and Pattern Recognition (CVPR) is known as the premier conference in the field of computer vision, pattern recognition, and related fields. CVPR, which is held annually in the U.S., has on average around 20\% acceptance rate for submitted papers, making it a very difficult conference to hurdle, and pointing to the high quality of the accepted papers. Also, CVPR is also known to comprehensively cover papers in the different fields in computer vision and pattern recognition. A number of prominent international researchers and research groups choose their research themes after a comprehensive grasp of almost all papers presented in premier conferences and an understanding of research trends. We believe that the accuracy by which research themes are chosen can be improved by constantly being updated on cutting-edge technologies and discussing these new technology trends within the research groups as part of their regular activities. Further, a survey of papers presented in premier conferences is also an essential way to gather tools needed for research. We therefore believe that gaining an understanding of papers presented in premier conferences is the best method for authors to comprehend the latest trends in computer vision, pattern recognition, and related fields. As the first step of this endeavor, we undertook to read all the 602 papers accepted during the CVPR2015~\cite{1_1,1_2,1_3,1_4,1_5,1_6,1_7,1_8,1_9,1_10,1_11,1_12,1_13,1_14,1_15,1_16,1_17,1_18,1_19,1_20,1_21,1_22,1_23,1_24,1_25,1_26,1_27,1_28,1_29,1_30,1_31,1_32,1_33,1_34,1_35,1_36,1_37,1_38,1_39,1_40,1_41,1_42,1_43,1_44,1_45,1_46,1_47,1_48,1_49,1_50,1_51,1_52,1_53,1_54,1_55,1_56,1_57,1_58,1_59,1_60,1_61,1_62,1_63,1_64,1_65,1_66,1_67,1_68,1_69,1_70,1_71,1_72,1_73,1_74,1_75,1_76,1_77,1_78,1_79,1_80,1_81,1_82,1_83,1_84,1_85,1_86,1_87,1_88,1_89,1_90,1_91,1_92,1_93,1_94,1_95,1_96,1_97,1_98,1_99,1_100,1_101,1_102,1_103,1_104,1_105,1_106,1_107,1_108,1_109,1_110,1_111,1_112,1_113,1_114,1_115,1_116,1_117,1_118,1_119,1_120,1_121,2_1,2_2,2_3,2_4,2_5,2_6,2_7,2_8,2_9,2_10,2_11,2_12,2_13,2_14,2_15,2_16,2_17,2_18,2_19,2_20,2_21,2_22,2_23,2_24,2_25,2_26,2_27,2_28,2_29,2_30,2_31,2_32,2_33,2_34,2_35,2_36,2_37,2_38,2_39,2_40,2_41,2_42,2_43,2_44,2_45,2_46,2_47,2_48,2_49,2_50,2_51,2_52,2_53,2_54,2_55,2_56,2_57,2_58,2_59,2_60,2_61,2_62,2_63,2_64,2_65,2_66,2_67,2_68,2_69,2_70,2_71,2_72,2_73,2_74,2_75,2_76,2_77,2_78,2_79,2_80,2_81,2_82,2_83,2_84,2_85,2_86,2_87,2_88,2_89,2_90,2_91,2_92,2_93,2_94,2_95,2_96,2_97,2_98,2_99,2_100,2_101,2_102,2_103,2_104,2_105,2_106,2_107,2_108,2_109,2_110,2_111,2_112,2_113,2_114,2_115,2_116,2_117,2_118,2_119,2_120,3_1,3_2,3_3,3_4,3_5,3_6,3_7,3_8,3_9,3_10,3_11,3_12,3_13,3_14,3_15,3_16,3_17,3_18,3_19,3_20,3_21,3_22,3_23,3_24,3_25,3_26,3_27,3_28,3_29,3_30,3_31,3_32,3_33,3_34,3_35,3_36,3_37,3_38,3_39,3_40,3_41,3_42,3_43,3_44,3_45,3_46,3_47,3_48,3_49,3_50,3_51,3_52,3_53,3_54,3_55,3_56,3_57,3_58,3_59,3_60,3_61,3_62,3_63,3_64,3_65,3_66,3_67,3_68,3_69,3_70,3_71,3_72,3_73,3_74,3_75,3_76,3_77,3_78,3_79,3_80,3_81,3_82,3_83,3_84,3_85,3_86,3_87,3_88,3_89,3_90,3_91,3_92,3_93,3_94,3_95,3_96,3_97,3_98,3_99,3_100,3_101,3_102,3_103,3_104,3_105,3_106,3_107,3_108,3_109,3_110,3_111,3_112,3_113,3_114,3_115,3_116,3_117,3_118,3_119,3_120,4_1,4_2,4_3,4_4,4_5,4_6,4_7,4_8,4_9,4_10,4_11,4_12,4_13,4_14,4_15,4_16,4_17,4_18,4_19,4_20,4_21,4_22,4_23,4_24,4_25,4_26,4_27,4_28,4_29,4_30,4_31,4_32,4_33,4_34,4_35,4_36,4_37,4_38,4_39,4_40,4_41,4_42,4_43,4_44,4_45,4_46,4_47,4_48,4_49,4_50,4_51,4_52,4_53,4_54,4_55,4_56,4_57,4_58,4_59,4_60,4_61,4_62,4_63,4_64,4_65,4_66,4_67,4_68,4_69,4_70,4_71,4_72,4_73,4_74,4_75,4_76,4_77,4_78,4_79,4_80,4_81,4_82,4_83,4_84,4_85,4_86,4_87,4_88,4_89,4_90,4_91,4_92,4_93,4_94,4_95,4_96,4_97,4_98,4_99,4_100,4_101,4_102,4_103,4_104,4_105,4_106,4_107,4_108,4_109,4_110,4_111,4_112,4_113,4_114,4_115,4_116,4_117,4_118,4_119,4_120,5_1,5_2,5_3,5_4,5_5,5_6,5_7,5_8,5_9,5_10,5_11,5_12,5_13,5_14,5_15,5_16,5_17,5_18,5_19,5_20,5_21,5_22,5_23,5_24,5_25,5_26,5_27,5_28,5_29,5_30,5_31,5_32,5_33,5_34,5_35,5_36,5_37,5_38,5_39,5_40,5_41,5_42,5_43,5_44,5_45,5_46,5_47,5_48,5_49,5_50,5_51,5_52,5_53,5_54,5_55,5_56,5_57,5_58,5_59,5_60,5_61,5_62,5_63,5_64,5_65,5_66,5_67,5_68,5_69,5_70,5_71,5_72,5_73,5_74,5_75,5_76,5_77,5_78,5_79,5_80,5_81,5_82,5_83,5_84,5_85,5_86,5_87,5_88,5_89,5_90,5_91,5_92,5_93,5_94,5_95,5_96,5_97,5_98,5_99,5_100,5_101,5_102,5_103,5_104,5_105,5_106,5_107,5_108,5_109,5_110,5_111,5_112,5_113,5_114,5_115,5_116,5_117,5_118,5_119,5_120,5_121}. This review summarizes all the papers of CVPR2015 we read as the first project of cvpaper.challenge. In this paper, we will describe the characteristics of CVPR2015 and discuss the trends and leading methods used in three areas; namely, recognition, 3D, and imaging/image processing. Further, we will enumerate the proposed datasets and new research problems presented at the conference and propose the concept of ``DeepSurvey". Finally, we will give a summary and discuss future steps. We would like to stress, however, that this paper mainly focuses on a survey of the research trends, and does not cover the details of all the 602 papers, which are beyond the scope of this paper.

\begin{figure*}[t]
\begin{center}
\includegraphics[width=150mm]{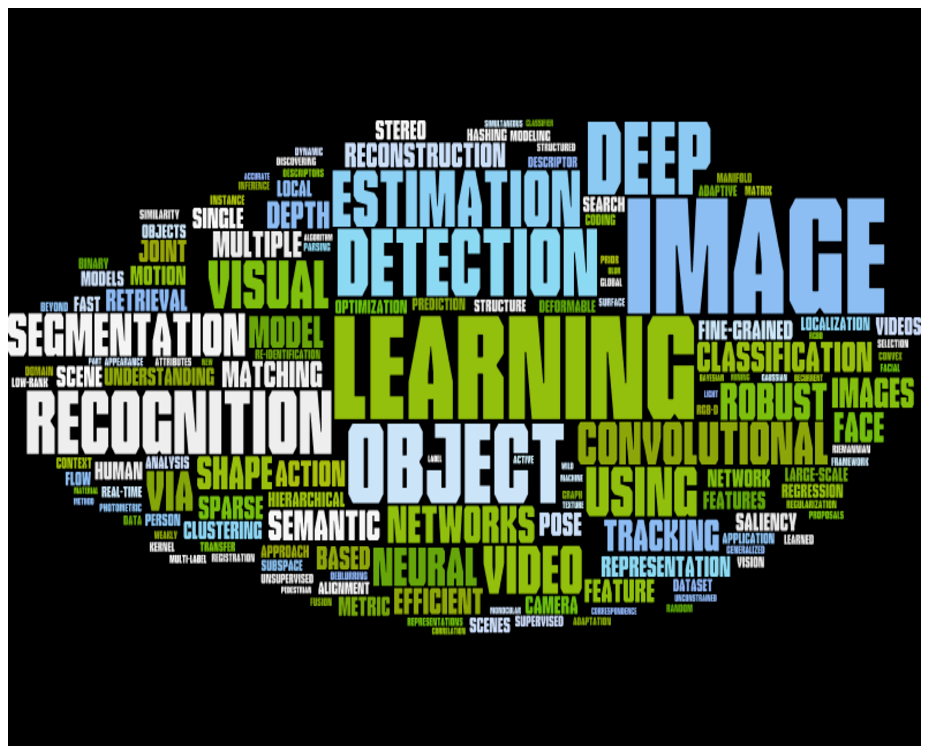}
\end{center}
\caption{Example of weighting and visualization based on titlesD}
\label{figure:wordle}
\end{figure*}

\section{Related initiatives and positioning of this project}

An example of a related initiative is the Japanese CHI Study Group that undertook to read all the papers presented at the ACM CHI Conference on Human Factors in Computing Systems, the top conference for user interfaces~\cite{CHI}. In 2015, the Study Group was held jointly in Tokyo and Hokkaido using a remote conference system to read within one day all the 485 papers presented at CHI2015. Speakers were assigned one session each and introduced one paper in approximately 30 seconds. The CHI Study Group started in 2006, and is on its 10th year in 2015. This initiative is a very effective way to grasp trends in the user interface domain, which is a very progressive field. It is noteworthy that several Japanese researchers have received the Best Paper and Honorable Mention awards at CHI2015.

Considering the rapid progress in the field, the authors focused on ``reading all the papers included in the international conference and summarizing trends through the project," as well as on listening to the introduction of the papers by project members. The CHI Study Group, therefore, serves its role in terms of covering all the papers. However, understanding the trends in a research field entails comprehensively reading all papers outside particular domains and holding discussions after reading the papers. We believe that trends can only be properly identified through discussions within the small group that undertook to read all the papers.

\section{Trends at CVPR2015}
This paper is mainly divided into three main areas; namely, (i) Recognition, (ii) 3D, and (iii) Imaging/Image processing. Before we proceed with discussing the details of each area, we will first explain the features and perspectives gleaned from the titles of the papers and of the papers selected for oral presentation.

\subsection{CVPR2015, as seen from the titles of the papers}
Figure~\ref{figure:wordle} shows a visualization of the titles of papers at CVPR2015 using Wordle~\cite{wordle}, a word-visualization service. We see that other than ``image" as the most frequent word, which is expected, we also see that the words ``deep," ``learning," ``object," and ``recognition" are very prominent. As can be seen from this word visualization and goes without saying, deep learning (DL) is a widely used tool in image recognition in researches presented at CVPR2015. Results of a search among CVPR2015 papers showed that the term was found in 250 out of the 602 papers. Although deep learning was mainly used for object recognition, since R-CNN~\cite{GirshickCVPR2014} was proposed, it also came to be used more for object detection. Also, with detection and recognition becoming more accurate, there are now more research initiatives to address semantic segmentation, which is an even more difficult problem. Meanwhile, the paper that received the Best Paper Award dealt with the method called DynamicFusion~\cite{1_38}, which pertains to real-time construction of high-resolution 3D models. The research was recognized for being able to successfully reconstruct non-rigid models in real time, in addition to being able to reconstruct in higher resolution than KinectFusion~\cite{NewcombeISMAR2011}. As shown by the prominence of the words ``reconstruction," ``depth," and ``shape," 3D research is also progressing at a steady pace. Also, deep learning is also being applied in stereo matching and 3D features, indicating the widening applications of deep learning. Even in imaging technologies, deblurring and sensing technologies are being updated, wherein examples of applications of feature extraction through deep learning were reported.

\subsection{CVPR2015 as seen from the oral presentations}
The 12 oral sessions are presented in CVPR2015 as below:
\begin{itemize}
 \item CNN Architectures
 \item Depth and 3D Surfaces
 \item Discovery and Dense Correspondences
 \item 3D Shape: Matching, Recognition, Reconstruction
 \item Images and Language
 \item Multiple View Geometry
 \item Segmentation in Images and Video
 \item 3D Models and Images
 \item Action and Event Recognition
 \item Computational Photography
 \item Learning and Matching Local Features
 \item Image and Video Processing and Restoration
\end{itemize}

\subsubsection{Recognition}
With CNN as the most widely used approach in deep learning, the theme of the first oral presentation was on CNN Architecture. First we would like to mention about GoogLeNet~\cite{1_1}, the winner of the ILSVRC2014 with a record 6.67\% top-5 error rate. GoogLeNet is a 22-layers CNN architecture, where convolutional structures are recursively connected to make a deep structure. Some presentations dealt with addressing the conventional problems in image recognition through deep learning, such as methods to implement multi-layer or multi-instance learning in order to achieve flexibility through shape change~\cite{1_43}, implementing optimization and repetition through Bayesian Optimization in the periphery of highly likely candidate regions in order to correct errors in object detection using R-CNN~\cite{1_28}, proposal of a robust expression to withstand positional invariability and deformations~\cite{1_108}, etc. Meanwhile, Nguyen \textit{et al.} automatically generated features that are mistakenly recognized by deep learning and showed that CNN features are not universal~\cite{1_47}. In Long \textit{et al.}’s segmentation method (FCN)~\cite{4_11}, inputting images results in an output wherein images segmented by pixel are outputted as fully connected layers. \cite{2_47} reports on the use of deep learning for morphing. A corresponding model of the chair is generated when the type of chair, camera viewpoint, and conversion parameters are inputted. There was also a paper on the output of multiple attributes from deep learning for crowded human environment~\cite{5_23}. It includes 94 attributes, and tags explaining “where,” “what kind of person,” “what is the person doing,” etc. are returned. Also, there was a research on visualization of features of each layer of CNN~\cite{5_81}, pointing to progress in the understanding of deep learning. A prominent session in the oral presentations for the area of recognition is “Images and Language” for image generation captions (image descriptions). In recent years, the level of research in this area has increased due to progress in research on recognition performance and natural language processing~\cite{3_98,3_101,3_44,3_55,3_83,3_13,3_106,2_40}. The “Action and Event Recognition” session formerly dealt with saliency and extension of human action recognition. Shu \textit{et al.} reported on taking aerial videos using drones (unmanned aerial vehicles (UAV)) to extract human lines of movement and recognize their group actions~\cite{5_14}. Fernando \textit{et al.} proposed VideoDarwin as a mechanism for capturing slowly changing actions in videos~\cite{5_102}. Ma \textit{et al.} expressed the hierarchy of each part of action recognition through integration of tree structures~\cite{5_63}. Khatoonabadi \textit{\textit{et al.}}~\cite{5_115} presented a method on saliency and Park \textit{\textit{et al.}} presented a method on social saliency prediction~\cite{5_36}. In~\cite{5_115}, they presented a method to achieve saliency and segmentation while reducing amount of information, based on image compression methods. Social saliency prediction~\cite{5_36} infers the area where attention is focused based on gaze directions of multiple persons.

\subsubsection{3D}
With DynamicFusion~\cite{1_38} at the head of the list, new methods on 3D were proposed. DynamicFusion is a method for conducting more precise 3D shape restoration in real time by chronologically integrating depth images obtained through Kinect and other methods. SUN RGB-D~\cite{1_62} was proposed as a large-scale data set that captures indoor space in three-dimension. Their dataset contains a total of 10,335 RGB-D images, and they presented several important issues on the topic. Research on occlusion using 3D models has also progressed. Xiang \textit{\textit{et al.}}~\cite{2_86} used 3D Voxel Patterns (3DVP) to carry out 3D detection and enabled detection of missing parts of objects where occlusion or defects have occurred, using a model-based method. “Reconstructing the World in Six Days” is an example of research on large-scale space~\cite{3_115}. They carried out 3D reconstruction through world-scale SfM of 100 million images of worldwide landmarks found in flickr. Hengel \textit{\textit{et al.}} used silhouette obtained from multiple images to carry out meaningful resolution of each part of the 3D model~\cite{1_96}. The 3D structure was realized through Block World~\cite{GuptaECCV2010}. Richter \textit{\textit{et al.}} proposed a method for discriminatively resolving Shape-from-Shading~\cite{2_2}. Albl \textit{\textit{et al.}} came up with a mechanism for properly operating, in global shutters as well as in local shutters, the perspective-n-point problem (PnP), which is considered important in SfM, inferring camera position, and calculating odometry~\cite{3_8}. Due to the problem of having an arbitrary rotation matrix, in their proposed method, they proposed an R6P algorithm to make more dense calculation of rotation matrix. Song \textit{\textit{et al.}} proposed a method to infer the 3D location of vehicles using monocular cameras~\cite{4_43}. Kulkarni \textit{\textit{et al.}} proposed Picture (Probabilistic Programming Language), which is a stochastic expression of 3D modeling, to enable expressing a more complicated generation model~\cite{4_114}. Wang \textit{\textit{et al.}} conducted 3D scene recognition in outdoor environment using GPS positional information as preliminary data~\cite{4_68}. CRF was applied to assign segments and their significance to 3D positional information. Barron \textit{\textit{et al.}} devised an optimized method to enable generation of effective stereo images~\cite{5_2}. Use of Defocus and Fast Bilateral Filter eliminates the need for calculating all corresponding points. Wang \textit{\textit{et al.}} devised a method for searching the 3D model from the 2D sketch~\cite{2_83}. A sketch image as seen from multiple perspectives is generated from the 3D model of one sample, and a 3D model is searched through comparison with inputted sketch image and presented to the user. Brubaker \textit{\textit{et al.}} carried out 3D molecular model reconstruction of high-resolution image from low-resolution image using electron cryomicroscopy~\cite{3_95}. Chin \textit{\textit{et al.}} realized improvement of robust matching such as RANSAC through optimization by A*search~\cite{3_21}.

\subsubsection{Image processing/imaging}
In regard to image processing and imaging, advances in research through new themes were seen. For example, Tanaka \textit{\textit{et al.}} presented their research resolution of paintings that are physically separated into multiple layers, such as pencil sketches or colored paintings~\cite{5_111}, enabling the extraction of even deeper components. \cite{5_73} presented the problem of finding an efficient border ownership (where the borderline is, whether an area is part of the background or foreground) in 2D images. The authors addressed the problem by using structural random forests (SRF) to differentiate borders. The problem regarding realizing photometric stereo under natural light rather than controlled light sources was also presented~\cite{5_8}. In order to apply photometric stereo in outdoor environment, the authors assumed a hemispherical experimental space and used GPS timestamp as preliminary information, and separately carried out light source estimation of sunlight. There were several proposals regarding the problem of inferring depth images from input images and videos, as well as a paper on simultaneous solution for image correction from fogged images and for inference of depth images~\cite{5_59}. Research on super-resolution was also included in the oral presentations~\cite{5_82}. The authors used self-similarity based super-resolution, and at the same time carried out inference of affine transformation parameters and localized shape variations.

\subsection{CVPR2015 as seen by area of study}
In the previous section we looked at CVPR2015 based on the titles and papers selected for oral presentation. In this section we will enumerate papers in more detail by area of study. Here we will comprehend the current trends in the field of computer vision by looking at all papers, regardless of whether they were presented orally or as posters.

\subsubsection{Recognition}

\textbf{Deep learning architecture.} We will cite papers that discuss the overall structure, as well as those that deal with problem-based structures, parameter adjustments, and architecture evaluation. Two examples of papers that discuss overall structure are on GoogLeNet~\cite{1_1} and DeepID-Net~\cite{3_20}. DeepID-Net uses Deformation Constrained (Def) pooling as alternative to max pooling and average pooling in order to improve expressiveness against changes in shape and position, as in DPM~\cite{FelzenszwalbPAMI2010}, contributing to improvement of accuracy in object detection. There were also many examples of attempts to carry out improvements under the framework of existing CNN methods~\cite{1_48,4_44,1_93}. Wan \textit{\textit{et al.}} combined the advantages of DPM and CNN and, further, implemented Non-maximum Suppression (NMS) in order to correct effects of positional discrepancies~\cite{1_93}. DPM is a method for preserving parts and position in latent variables, while CNN has the advantage of being able to automatically learn features that are useful for object recognition. Other papers dealt with the characteristics of CNN~\cite{1_47,1_43,1_108}, increasing speed of learning~\cite{1_88}, initiatives to search for parameters~\cite{5_99}, and visualization of features~\cite{5_81}. Lenc \textit{\textit{et al.}} carried out robust CNN feature expression to address image rotation by implementing a transformation layer for geometric transformation of convoluted features~\cite{1_108}. Liu \textit{\textit{et al.}} succeeded in reducing computational complexity and CNN calculation time by implementing sparse representation to address convolution~\cite{1_88}. They succeeded in significantly reducing calculation time by sparsing of kernels computed at every convolution, and improved calculation to enable operation even on a CPU. He \textit{\textit{et al.}} studied depth of structure, filter size, stride, and other trade-offs pertaining to CCN architectural parameters ~\cite{5_99}, and showed that depth is important. Other papers dealt with improvement of convolution layers~\cite{4_4}, method to calculate similarity of patches~\cite{3_114,4_110}, and research on morphing under the CNN framework~\cite{2_47}. Liang \textit{\textit{et al.}} claimed that better features can be obtained if CNN convolution frameworks are recursively convoluted~\cite{4_4}. This structure is called Recurrent Convolutional Layter (RCL). In MatchNet, architecture is configured for the purpose of measuring similarity between patches, and is partitioned to a network for generating features through pooling and convolution of four layers and a network for evaluating similarity through total combination of three layers~\cite{3_114}. Zagoruyko \textit{\textit{et al.}} also discussed a framework for calculating patch similarities in CNN~\cite{4_110}. They extracted the features based on convolutions of paired patches and calculated similarity in the later layers.

\textbf{Human recognition.} We will introduce papers in Human Recognition by dividing them into Face Recognition，Pedestrian Detection，Human Tracking，Pose Estimation，Action Recognition，Event Recognition，Crowd Analysis，Egocentric Vision，and Person Re-identification. 

First, in \textbf{face recognition}, FaceNet was presented as a system for handling high-precision recognition~\cite{1_89}. DeepFace, which has been recently proposed in 2014~\cite{TaigmanCVPR2014}, brought about significant improvements in accuracy, but FaceNet has achieved an even higher accuracy than DeepFace. Sun \textit{\textit{et al.}} improved their conventional face recognition, DeepNet~\cite{SunNIPS2014}, and applied features extracted from early convolution layers to improve face recognition accuracy particularly of face profiles and occlusions~\cite{3_73}.

In \textbf{pedestrian detection}, Tian \textit{\textit{et al.}} were able to improve accuracy by combining CNN features and attributes for detection of pedestrians~\cite{5_69}. They accomplished this by including other attributes, such as positional relationships between pedestrians and environment, as well as learning of pedestrians and backgrounds. Honsang \textit{\textit{et al.}} implemented evaluation of features using CNN to carry out pedestrian detection~\cite{4_80}.

In \textbf{pose estimation}, a research on marker-less motion capture using CNN features was presented~\cite{4_51}. For practical use, it is possible to significantly reduce installation costs if estimation can be implemented through maker-less MoCap using 23 cameras.

In \textbf{human tracking}, there were reports featuring more advanced methods. Milan \textit{\textit{et al.}} were able to simultaneously carry out tasks of chronological area estimation and positioning by using Superfixel and CRF~\cite{5_104}. They established a method for combining low- and high-level information and finely dividing background and foreground. A method for carrying out accurate tracking of multiple objects using Target Identity-aware Network Flow (TINF), which probabilistically resolves network nodes, was also presented~\cite{2_4}. The method constructs the optimum network using graph theory and carries out optimization through Lagrangian relaxation optimization. In action recognition, Gkioxari \textit{\textit{et al.}} used R-CNN~\cite{GirshickCVPR2014} as basis for proposing a mechanism for recognizing actions, including position of the human subjects~\cite{1_83}. In order to extract the action area, candidate areas where extracted from an assembly of optical flows to extract CNN-based features. And in order to extract features from chronological actions, convolution was implemented for chronological images that stored optical flows and RGB visible images. To improve accuracy, researchers proposed a method based on Dense Trajectories (DT)~\cite{WangCVPR2011,WangICCV2013} and on TDD, an action descriptor that combines CNN features~\cite{5_102}. In regard to the DT-based method, researchers adopted HOG, HOF, and MBH to accurately recognize actions, as well as applied CNN features to action recognition through normalization of the feature map. Lan \textit{\textit{et al.}} proposed Multi-skip Feature Stacking (MIFS), a method for extract features by configuring multiple gradations to a chronological offset~\cite{1_23}.

In \textbf{event recognition}, architecture specialized for event recognition called Deep Event Network (DevNet) was proposed~\cite{3_38}. The system enabled extracting not only pre-defined events, but also clues for important chronological events. Xiong \textit{\textit{et al.}} carried out recognition of complex events by combining multiple identification results and factors for still images and combined CNN features and results of object/human/face detection results to recognize events~\cite{2_54}. Shu \textit{\textit{et al.}} carried out event recognition from aerial images taken using unmanned aerial vehicles (UAV)~\cite{5_14}. They proposed a Space-time AND-OR Graph to analyze various clues from images from drones, such as positional adjustment of images containing egomotion, group action recognition, and human interaction.

In \textbf{crowd analysis}, a mechanism that allows cross-scene crowd counting was proposed~\cite{1_91}. They used a CNN model that allows switching the crowd density map and human count model. Although these two models are different, they are correlated and complement each other’s accuracy. Yi \textit{\textit{et al.}} analyzed crowd models from videos taken from surveillance cameras and measured routine pedestrian path directions~\cite{4_17}. They predicted crowd attributes and pedestrian destinations and enabled detection of abnormal actions as well as prediction of paths taken to reach destinations. 

A method for editing one’s own videos taken using \textbf{egocentric vision} was also proposed~\cite{5_109}. Research to solve face recognition problems, such as recognition of severely occluded faces and small and far faces in images, has progressed. Huang \textit{\textit{et al.}} proposed a hand region segmentation method for egocentric vision to determine what tasks the person taking the video is performing~\cite{1_73}. 

\textbf{Person re-identification} deals with the problem of personal authentication between different cameras for surveillance and other cameras. Shi \textit{\textit{et al.}} inferred semantic attributes regarding humans and clothing at the patch level, and applied them in person re-identification~\cite{4_92}. They obtained clothing and other external appearance features and were able to improve expressivity by using attributes. Chen \textit{\textit{et al.}} carried Multiple Similarity Function Learning using PCA compression color and texture features from images with segregated regions~\cite{2_50}. Zheng \textit{\textit{et al.}} evaluated effectiveness of features and enabled feature integration needed for Re-ID using Late Fusion~\cite{2_69}. Person re-identification using low-resolution images was also addressed~\cite{1_76}. Generally, images from surveillance cameras are of poor quality, and to address this, Jing \textit{\textit{et al.}} carried out super-resolution to propose a mechanism for improving performance even for low-resolution images. Neural network architecture to improve robustness against feature variations between cameras was also proposed~\cite{4_62}. Given a pair of images as input, the authors used the difference of activation functions extracted from each patch after convolution and pooling as features for recognition.

\textbf{Object recognition and detection.} The problem of recognizing objects appearing in images is currently an intensively studied area. This section also deals with object detection that includes recognition of position, scene recognition, search of hashed images, as well as fine-grained image recognition. Papers on object recognition have dramatically increased after AlexNet was proposed~\cite{KrizhevskyNIPS2012} at ILSVRC2012, and object recognition has also been applied to scene recognition and other problems. Research on object detection expanded after the proposal of R-CNN~\cite{GirshickCVPR2014}. These trends are clearly evident in CVPR2015.

A study was conducted to improve accuracy and streamline recognition by carrying out selection of CNN factors~\cite{1_106}. Association Rules~\cite{AgrawalSIGMOD1993} widely used in the data mining field were applied, and only features that are useful for identification were selected as a subset from among the CNN feature space. In object detection, there were many researches addressing the problem of “inaccurate localization,” which is one of the vulnerabilities of R-CNN. As previously mentioned, Zhang \textit{\textit{et al.}} proposed a method for optimization to correct inaccurate localization in R-CNN to address this vulnerability~\cite{1_28}. Tsai \textit{\textit{et al.}} considered the diversity of internal changes and variations of objects for detection, and compensated for inaccurate localization by improving feature pooling~\cite{1_80}. Oquab \textit{\textit{et al.}} used weakly supervised learning to investigate solutions for discrimination and localization of objects based only on labeling of image levels~\cite{1_75}. Fine-grained image discrimination is a problem that entails more detailed classification of objects, such as dog breeds or vehicle types. Due to high visual similarity of objects, such detailed classification is very difficult to carry out. It was found that adaptively extracting features useful for discrimination by dividing images into parts and extracting features only from particular regions is an effective method~\cite{FINEGRAINED}. Using CNN architecture, Xiao \textit{\textit{et al.}} extracted candidate patches from major categories (e.g. dog, bird) and detailed categories (e.g. fine classification of dogs and birds) in a layered structure, and simultaneously implemented feature selection and discrimination~\cite{1_92}. Xie \textit{\textit{et al.}} carried out learning by applying multitask learning in multiple structured classes as well as in limited –task data extensions~\cite{3_46}. They succeeded in simultaneously learning relationships through multitask learning of major and minor classifications. Lin \textit{\textit{et al.}}~\cite{2_61} proposed Deep Localization, Alignment and Classification (DeepLAC) as a mechanism to correct changes in regional position and angles, which is needed for fine-grained image recognition, within the back-propagation algorithm framework.

\textbf{Segmentation.} Segmentation requires implementing object recognition at the pixel level, making it a difficult procedure in terms of distinguishing borders between foreground and background. The number of papers dealing with semantic segmentation, which deals with assignment of meaning to segmentation areas, has increased.

Hariharan \textit{\textit{et al.}} demonstrated the increase in accuracy of semantic segregation by using features extracted in the middle layers, not only from the fully connected layer, in regard to CNN architecture~\cite{1_49}. In particular, they used the 2nd pooling layer, the 4th convolution layer, and the 7th fully connected layer, and by combining these they were able to simultaneously implement low-, mid-, and high-level feature expression. In saliency-based segmentation, a method was proposed for extracting multi-scale CNN features~\cite{5_110}. Itti \textit{\textit{et al.}}’s saliency model is well known~\cite{IttiVR2000}, and, although they conducted multi-scale calculations, Li \textit{\textit{et al.}} extracted saliency and applied it segmentation by replacing CNN features. Although it overlaps with 3D reconstruction, we would like to mention here that Martinovic \textit{\textit{et al.}} proposed research for implementing semantic segmentation of 3D urban models~\cite{5_1}.

\textbf{Data generation.} Data generation is an important issue in addressing recognition problems. In this section we will cite papers on data collection and selection. Hattori \textit{\textit{et al.}} generated learning images for pedestrian detection~\cite{4_52}. They conducted learning of 36 types of pedestrians, various kinds of walking, and occlusion patterns using CG. Russakovsky \textit{\textit{et al.}} cited an annotation method leveraging crowdsourcing, in order to efficiently and accurately detect objects~\cite{2_110}. The method deals with the usability and accuracy of labeling and is aimed at minimizing human annotation costs, wherein machines and humans interactively carry out annotation based on results from baseline recognition equipment. Xiao \textit{\textit{et al.}} discussed a framework for efficient labeling and learning, in an effort to reduce annotation operations for massive data~\cite{3_51}.

\subsubsection{3-Dimension}

There were also many examples of applications of CNN even for 3D object recognition. Fang \textit{et al.} proposed Deep Shape Descriptor (DeepSD) as a method for expressing 3D shapes~\cite{3_11}. They proposed a robust 3D feature that can handle structural variations in shape, noise, and shapes that include three-dimensional incompleteness. Xie \textit{et al.} proposed DeepShape, a CNN feature to address problems in 3D object matching and retrieval~\cite{2_18}. They used a shape descriptor based on an auto-encoder to search 3D shapes. Abdelrahman \textit{et al.} proposed a 3D non-rigid texture descriptor based on Weighted Heat Kernel Signature (W-HKS)~\cite{1_21}. There was also a proposal for a mechanism to extract information useful for recognition even from a limited learning sample using Deep Boltzmann Machine (DBM) and design of object recognition features through RGB-D~\cite{3_86}. They proposed an effective descriptor even for complex 3D objects by combining geometric shape information as well as color information.

In RGB-D input, a problem was reported in giving tasks, such as 3D recognition and inferring positions that can be grasped by robots, in complex indoor environment~\cite{5_17}. Superfixel was applied as a preliminary processing step, and recognition of cuboid models and spatial smoothing through Conditional Random Fields (CRF) was carried out. Matsuo \textit{et al.} also proposed a method for enhancing depth images (particularly planes) by combining low-resolution depth images and high-resolution RGB images~\cite{4_26}. They adjusted position and connection of tangent planes in 3D space and used JBU filter to reconstruct rough surfaces. Gupta \textit{et al.} conducted research on extracting object position and 3D segmentation results from RGB-D image input~\cite{5_31}. They expressed object features through learning by CNN of surface normal line images. They then roughly estimated object pose based on 3-layered CNN and inferred detailed object pose and segment by comparison with the 3D model.

\subsubsection{Image processing/imaging}

CNN was also used for blur removal~\cite{1_84}. Non-uniform motion blurs arising from shaking of camera, etc. were corrected through learning of blurred/non-blurred patch pairs. There was also a research on fusion of multiple kernels~\cite{1_41}. The authors adopted a method using kernels for fusing multiple deblurring methods in order to develop a more advanced blur removal method. By using Gaussian Conditional Random Fields (GCRF), they were able to carry out kernel fusion based on learning. Eriksson \textit{et al.} proposed a method for noise removal that takes sparsity into consideration~\cite{4_2}. To solve the k-support norm optimization and normalization problem, Eriksson \textit{et al.} carried optimization by considering this problem as the minimum convex set that includes the set given as Convex Envelopes. Research on blur removal for videos was also reported~\cite{4_76}. There are two methods for blur removal for videos. One is by independently removing blur within the frames and splicing the frames together. The other is by inferring camera motion between frames. Zhang \textit{et al.} combined these two methods.

In regard to the problem of super-resolution, a method using Self-Similarity based Super-Resolution was reported~\cite{5_82}. The method simultaneously infers affine transformations and localized shape variations. Comparison with external/internal dictionaries enabled mapping to clear images. A method using a reference dictionary that accommodates shape variations was also reported to address the super-resolution problem for single images~\cite{5_106}. Gradient Ridge Image processing was performed as a preliminary processing step, and resolution was enhanced through matching with the dictionary. Schulter \textit{et al.} solved the single-image super-resolution problem as a linear regression problem using Random Forests~\cite{4_49}. 

A method for inferring shadow regions using CNN was reported for basic algorithms in image processing~\cite{2_104}. Shen \textit{et al.} also proposed DeepContour, which is a CNN architecture for contour detection~\cite{4_70}. DeepContour involves learning contour/non-contour regions and is composed of a 6-layered architecture (four convolution layers and two fully connected layers). DeepEdge was also proposed as an application of CNN architecture for edge detection~\cite{4_113}. DeepEdge carries out more accurate edge detection by using higher-level features. Experimentally, they were able to show that unlike CannyEdge, where there was noise contamination, DeepEdge was able to better remove backgrounds as well as extract edges from objects. Teo \textit{et al.} also proposed a method for effectively extracting borderlines in 2D images~\cite{5_73}. By using Structural Random Forests (SRF), they were able to rapidly determine where the borders are, and whether the area belongs to the background or the foreground. A research on the application of Linear Spectral Clustering (LSC) to Superpixel methods was also presented~\cite{2_27}.

In \textbf{device research}, a hyperspectral camera that can acquire chronological images was proposed~\cite{5_54}. Sequences of multiple hyperspectral cameras were alternately complemented, and image reconstruction based on dictionary learning was conducted, in order to obtain clear images even at high-speed (100 fps) observation. Ti \textit{et al.} developed a ToF sensor using a monocular camera and LED~\cite{4_108}. They developed the ToF sensor by attaching a total of four LEDs to the upper, lower, right, and left sides of the camera and capturing the reflection of LED light using the camera. To improve accuracy of ToF cameras, Naik \textit{et al.} resolved the problem of Multipath Interference (MPI), where multiple optical reflections appear and are mixed up in the pixel~\cite{1_9}. MPI also occurs in natural scenes, such as in an environment where multiple reflected lights occur or reflected light is diffused. These reflections were divided into Phase and Amplitude, both directly and globally, in order to reduce depth image errors due to MPI. Ye \textit{et al.} proposed an enhanced Kinect sensor by attaching Ultrasonic Sensor to Kinect~\cite{5_48}. They inferred the plane by applying Bayesian Network to the inference point obtained through the Ultrasonic Sensor.

\subsubsection{Datasets}

In this section, we will discuss new research problems as well as research on datasets.

\textbf{Datasets.} An example of a dataset is the SUN RGB-D, an expansion of SUNdatabase (which is a problem in scene recognition mentioned in the previous section) to RGB-D~\cite{1_62}. It is an attempt to expand the data set into more advanced scene recognition, such as segmentation and detection of objects within scenes, other than merely for recognition. A similar research problem is on the dataset for estimating indoor layout proposed by Liu et al~\cite{4_9}. The dataset for indoor environment included information on the entire room, walls, doors, windows, and their positional information. A research for outputting detailed explanations of medical images was also reported~\cite{1_119}. This research problem pertained to outputting sentence descriptions from an input of medical images. Detailed explanations of symptoms are generated by learning in pairs the actual medical images and the corresponding medical examination results. There was also an attempt to increase recognition capability by creating a much larger-scale dataset in the field of fine-grained recognition~\cite{1_65}. NABirds is a dataset for fine-grained recognition of birds, the scope of which was expanded by increasing the number of classes. There was also a report on a dataset for categorizing cars~\cite{4_69}. The study provided data for fine-grained classification of cars, which previously were only categorized into the class called “cars.”

There was also a study on creating data for detection of pedestrians through the use of images containing a higher amount of information. Hwang \textit{et al.} used a hyperspectral camera to acquire richer image information in order to improve detection of pedestrians at nighttime as well as daytime~\cite{1_113}. A dataset was also proposed for analyzing each person in a crowd by focusing on the spectators rather than on the sport itself~\cite{2_101}. They analyzed individual reactions of persons in a crowd, categorized crowds, and determined the type of spectators. In regard to pedestrian detection, a dataset was proposed for estimating gender, age, weight, clothing, etc., of pedestrians as well their location~\cite{5_113}. This dataset is intended for fine-grained recognition of persons. Thus, there was more focus on addressing fine-grained detection of pedestrians. There is more research being conducted on generation of image descriptions, with one oral session devoted to the topic. In particular, Rohrbach \textit{et al.} proposed a dataset for movie description~\cite{3_106}. For action recognition datasets, Heilbron \textit{et al.} published a dataset called ActivityNet~\cite{1_105}, which is a large-scale dataset similar to ImageNet and includes a significantly greater amount of data and action variations. The dataset includes 203 trimmed data classes and 137 untrimmed classes, for a total of 849 video hours. Also in action recognition, Xu \textit{et al.} proposed a dataset that maps attributes in advance to actors and actions~\cite{3_5}. 

\begin{figure*}[t]
\begin{center}
\includegraphics[width=150mm]{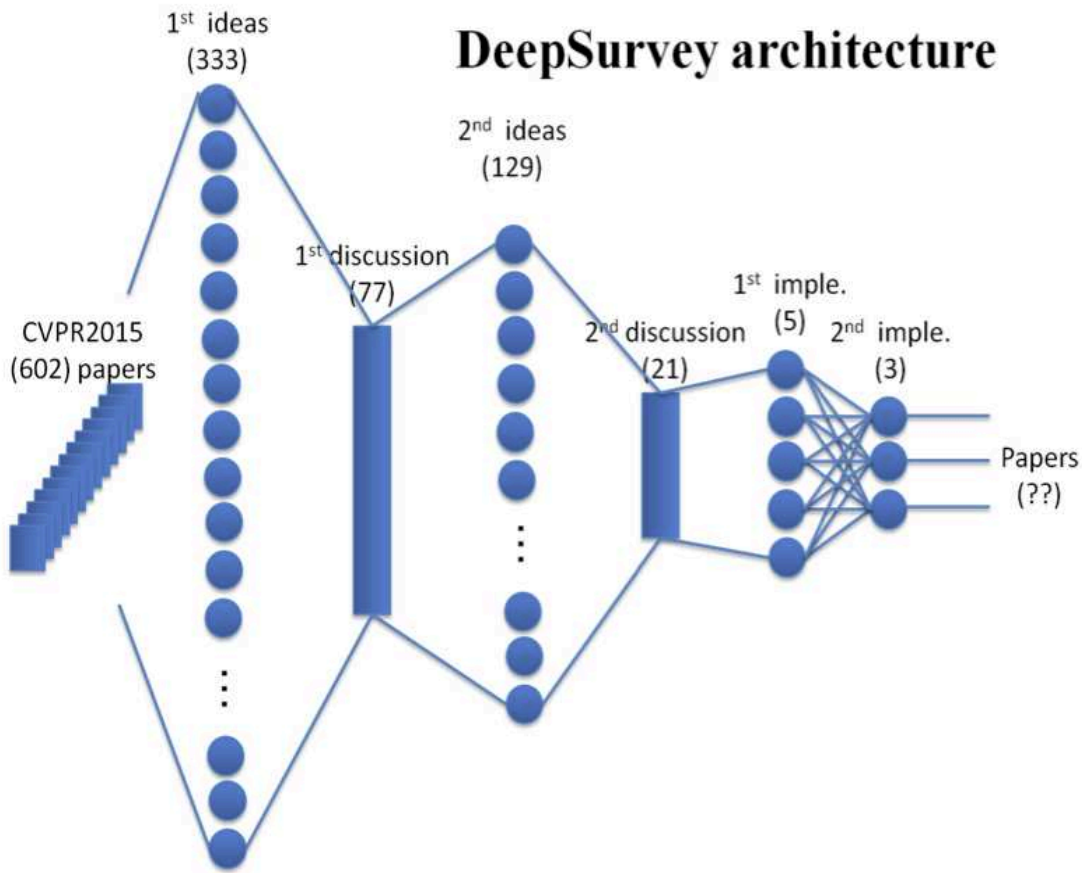}
\end{center}
\caption{DeepSurvey architecture: ( ) shows the actual number of papers and the number of ideas and implementations.}
\label{figure:deepsurvey}
\end{figure*}

\textbf{New research problems.} Here we introduce new research problems proposed at CVPR2015. Lin \textit{et al.} proposed the research problem of identifying locations of aerial images using images taken on the streets as query~\cite{5_61}. Although ground images and aerial images are completely different in nature, the authors presented a possible approach to the problem by proposing Where-CNN. Akhter \textit{et al.} conducted estimation of 3D human pose from 2D joint angles, and by adding a joint angle limit they were able to add a process for reducing poses with inscrutable motion~\cite{2_37}. Peng \textit{et al.} proposed two new aspects on human emotions predicted from images~\cite{1_94}. There was a paper on detecting persons or animals in a “best” relationship, i.e., with a high co-occurrence relationship with another person or animal based on Best-Buddies Similarity~\cite{2_99}. The authors proposed a method based on template matching to visualize the co-occurrence relationship. There was also a paper that addressed the problem of identifying very important people (VIP) within a group~\cite{5_45}. The authors used im2text to solve the problem by classifying level of importance of images and texts. Traditional machine learning methods map input and output vectors as pairs, but Wang \textit{et al.} assigned hidden information to images to further improve flexibility~\cite{5_57}. On the basis of this concept, they proposed that hidden information be handled as features or second objective functions. Zhang \textit{et al.} proposed a method to address the problem of counting items in an image as well as finding saliency from images~\cite{4_77}. They claimed that it can be used for egocentric lifelogs and image thumbnails. Not only in sensing, but there will also be a need to carry out person recognition in next-generation camera images whose resolution has been lowered for security and privacy protection. This problem is addressed by Pittaluga \textit{et al.} by carrying out face and pose recognition that can handle low-resolution images and resist changes in light source, proposing the method to be used for privacy protection~\cite{1_35}. There was also a proposal on object recognition that takes into consideration what kind of tasks are completed using particular tools~\cite{3_69}. The authors constructed 3D models of objects using 3D sensors and inferred the position by which the person carries the object based on joint angle, and measured how the task is being carried out. Measurement was made not only on 2D and 3D images, but they also calculated the impulse strength using voice data. Handling of the tool was inferred based on joint angle trajectory. There was also a proposal for inferring what a store is selling based on the storefront image~\cite{2_64}. Streetview images were used to extract characters through OCR, and ontology from those characters was used to classify stores according to business category.

\section{DeepSurvey}

We are proposing DeepSurvey (see Figure 2) as a mechanism for the systematization of knowledge, the generation of ideas, and as well as the writing of papers (specially for new research problems) based on an extensive reading of papers. DeepSurvey architecture is devised based on DeepLearning, which has flourished in recent years, and is composed of the following elements:

\begin{itemize}
    \item Input: Input the papers read (knowledge)
    \item 1st ideas: Individually generate ideas (from knowledge to ideas)
    \item 1st discussion: Group discussion (consolidation of ideas)
    \item 2nd ideas: Generate more ideas based on consolidated ideas
    \item 2nd discussion: Further refinement of ideas
    \item 1st implementation: Pick-up and hackathon
    \item 2nd implementation: Full-scale implementation and experiment
    \item Output: Paper
\end{itemize}

In comparison with general Convolutional Neural Networks (CNN)~\cite{LeCun1998}, “ideas” can be replaced with “convolution layer,” “discussion” with “pooling,” and “implementation” with “fully connected layer” to make it easier to understand. In “pooling” (discussion), multiple ideas are collected and good ideas are inputted as they are to the next layer, thus, it is closely similar to Lp pooling, which simultaneously possesses characteristics of max pooling and average pooling. The strategy is to repeat generation of ideas and discussion, and proceed to implementation once ideas have taken shape. The current counting of layers include convolutional layers and fully connected layers, thus, the architecture is a four-layer configuration.  

The most important feature of this architecture is the method for ``becoming a part of the neuron." Under this framework, since the entire group works as one neural network architecture in real, rather than in virtual space, the group is able to write papers as the final output. (Thankfully, we got first output of DeepSurvey~\cite{Kataokaarxiv2016} which includes a conceptual subject integrating semantic segmentation into change detection.) It is also characterized by project members actually doing the thinking, reading, and writing of papers to enable them to grow, wherein the network itself grows and matures.

For 2015, there was little time left for implementation and writing of papers, but we would like to write a more refined paper in the next year as well as be able to propose new research problems. Recently, since the structure of the architecture is also becoming deeper (VGGNet~\cite{SimonyanICLR2015}: 16/19 layers; ResNet~\cite{HeCVPR2016} 50/101/152 layers), going forward, we would like to generate more ideas, hold more discussions, and produce more refined ideas, research problems, and papers.

\section{Summary and future trends}

In this survey we comprehensively read papers presented at CVPR2015 to gain an understanding of the trends in computer vision. Further, we devised DeepSurvey as a mechanism to generate ideas from knowledge and eventually write a paper. We divided the papers into three areas; namely, recognition, 3D, and imaging/image processing, and sought to identify new research areas, as a means to expand the limits of the field. Here we are proposing DeepSurvey, and, going forward, we have started addressing some of its problems.

The authors are sorting out the current issues and believe that conducting surveys that include a study of technologies is essential also for identifying the next research problems. Further, there is a need to gain the ability to view the field from a wider perspective aside from actually testing the survey results to better understand the issues. We hope that this initiative would serve as a useful step towards that end.


\end{document}